\documentclass{article}

\usepackage[preprint]{neurips_2024}
\usepackage{caption}
\usepackage{svg}

\usepackage[utf8]{inputenc} 
\usepackage[T1]{fontenc}    
\usepackage[colorlinks = true,
            linkcolor = blue,
            urlcolor  = blue,
            citecolor = blue,
            anchorcolor = blue]{hyperref}   
\usepackage{url}            
\usepackage{booktabs}       
\usepackage{amsfonts}       
\usepackage{nicefrac}       
\usepackage{microtype}      
\usepackage{graphicx}
\usepackage{amsmath}
\usepackage{array}
\usepackage{multirow}
\usepackage{subcaption}
\usepackage{siunitx}
\usepackage{xcolor} 
\usepackage{tcolorbox} 
\usepackage{enumitem}

\title{Low-Bit Quantization Favors Undertrained LLMs:\\Scaling Laws for Quantized LLMs with 100T Training Tokens}

\author{Xu Ouyang$^{1,2}$\thanks{Work done while interning at Tencent AI Lab Seattle.}~~~~~Tao Ge$^2$\thanks{Corresponding author}~~~~~Thomas Hartvigsen$^1$~~~~Zhisong Zhang$^2$~~~~Haitao Mi$^2$~~~~Dong Yu$^2$ \\
$^1$University of Virginia ~~~~~~~~~~~~~~~$^2$Tencent AI Lab Seattle~~~~~ \\
\texttt{ftp8nr@virginia.edu}~~~~~~~~\texttt{getao@global.tencent.com}
}

\begin{document}
\maketitle
\linespread{1.1}
\begin{abstract}
We reveal that low-bit quantization favors undertrained large language models (LLMs) by observing that models with larger sizes or fewer training tokens experience less quantization-induced degradation (QiD) when applying low-bit quantization, whereas smaller models with extensive training tokens suffer significant QiD. To gain deeper insights into this trend, we study over 1500 quantized LLM checkpoints of various sizes and at different training levels (undertrained or fully trained) in a controlled setting, deriving scaling laws for understanding the relationship between QiD and factors such as the number of training tokens, model size and bit width.

With the derived scaling laws, we propose a novel perspective that we can use QiD to measure an LLM's training levels and determine the number of training tokens required for fully training LLMs of various sizes. Moreover, we use the scaling laws to predict the quantization performance of different-sized LLMs trained with \textbf{\textcolor{red}{100 trillion}} tokens. Our projection shows that the low-bit quantization performance of future models, which are expected to be trained with over 100 trillion tokens, may NOT be desirable. This poses a potential challenge for low-bit quantization in the future and highlights the need for awareness of a model's training level when evaluating low-bit quantization research. To facilitate future research on this problem, we release all the 1500+ quantized checkpoints used in this work at \url{https://huggingface.co/Xu-Ouyang}.
\end{abstract}\vspace{-0.2cm}

\begin{figure}[h!]
\centering
    \includegraphics[width=1\textwidth{}]{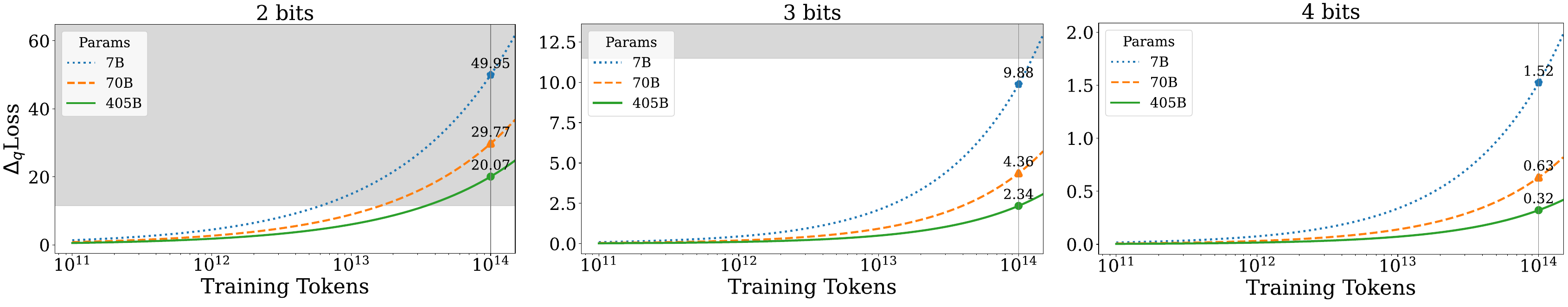}\vspace{-0.2cm}
    \caption{Scaling laws for predicting \textbf{Q}uantization-\textbf{i}nduced \textbf{D}egradation (QiD, denoted as $\Delta_qLoss$) in 7B, 70B, and 405B models trained on up to 100 trillion ($10^{14}$) tokens. While low-bit quantization yields acceptable QiD for undertrained LLMs (trained with $\leq 10^{12}$ tokens), it is predicted to become undesirable when applied to fully trained LLMs (e.g., trained with 100 trillion tokens, a milestone expected to be reached in the next few years), particularly for smaller models. Note that the \colorbox{lightgray}{gray areas} in this figure indicate levels of QiD that degrade the model's predictions to a level worse than random guessing.}\vspace{-0.2cm}
    \label{fig:100T}
\end{figure}

\section{Introduction}

Quantization \citep{jacob2018quantization,krishnamoorthi2018quantizing,banner2019post,frantar2022gptq,shen2024agile,lin2024awq,zhang2024lqer} is one of the most popular techniques for efficiently deploying large language models (LLMs) by reducing the model’s disk size, memory footprint, and improving inference efficiency through lower precision weights and activations. As model sizes have continued to grow over the past years, researchers have moved beyond conventional 8-bit quantization \citep{zafrir2019q8bert, dettmers2022gpt3, zhong2024erq} and begun exploring even lower bit width \citep{bai2020binarybert, zhang2020ternarybert,wang2023bitnet,liu2023qllm,egiazarian2024extreme, liu2024kivi, huang2024billm}, sparking a surge of research interest in low-bit quantization.

\begin{figure}[t]
\centering
    \includegraphics[width=0.98\textwidth{}]{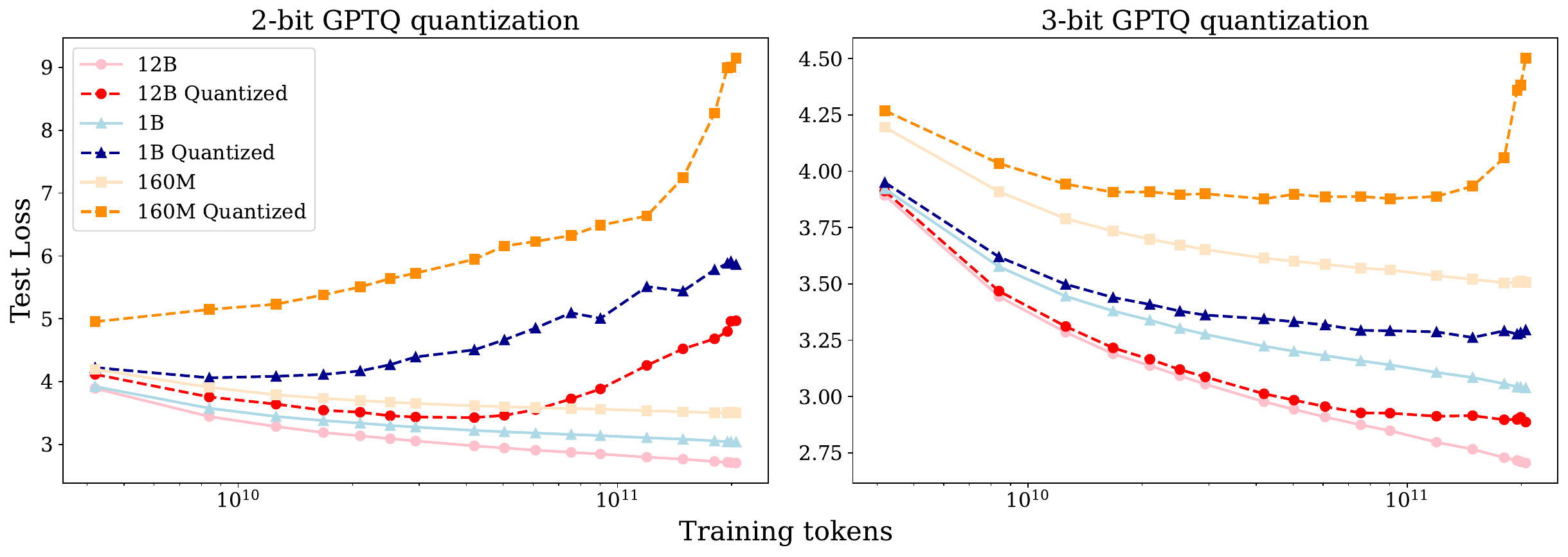}\vspace{-0.2cm}
    \caption{Performance of LLMs after low-bit quantization at different sizes and training levels. It is obvious that the models which are smaller or trained with more tokens suffer from greater quantization-induced degradation.}\vspace{-0.2cm}
    \label{fig:intro_quant}
\end{figure}

While low-bit quantization works well on some LLM checkpoints with very little quantization-induced degradation (QiD), we have observed that these checkpoints are typically with either larger model sizes or fewer training tokens. In contrast, smaller models or those trained with much more tokens tend to suffer from significant QiD when low-bit quantization is applied. As shown in Figure \ref{fig:intro_quant}(right), 3-bit quantization results in negligible QiD for a 12 billion parameter LLM up to $10^{11}$ training tokens, but beyond this point, QiD begins to become pronounced; For smaller models (e.g., 160 million and 1 billion parameters), QiD degradation occurs much earlier and is more severe. With even more extreme 2-bit quantization as shown in Figure \ref{fig:intro_quant}(left), the trend is similar, but QiD worsens sooner and more significantly. This observation suggests that low-bit quantization tends to favor undertrained LLMs and is less compatible with fully trained LLMs.

To gain deeper insights into this trend, we study over 1500 quantized LLM checkpoints of various sizes (ranging from 160M to 12B) and at different training levels\footnote{Training levels in this work refer to the extent to which an LLM has been trained (e.g., undertrained, fully trained, or overtrained), which are related to both the number of training tokens and the model size.} (trained with from 1B to 206B training tokens), analyzing the impact of low-bit quantization on them in a controlled setting. We derive scaling laws to model QiD with respect to the number of training tokens, model size, bit width. According to the derived scaling laws, we propose a novel perspective that we can use QiD to measure an LLM's training levels and determine the number of training tokens required for fully training an LLM given its size. Moreover, we use the scaling laws to predict the performance of different-sized LLMs with 100 trillion training tokens when applying low-bit quantization. Our projection shows that low-bit quantization of future models, which are expected to be trained with over 100 trillion tokens, may not be desirable, which indicates a potential challenge for low-bit quantization in the future and suggests that a model's training level should be considered in the evaluation of future low-bit quantization research.

The contributions of this work are threefold:
\begin{itemize}[left=5pt]
    \item We reveal that low-bit quantization favors undertrained LLMs but suffers from significant quantization-induced degradation (QiD) when applied to fully trained LLMs. This insight has been largely overlooked in previous low-bit quantization research: very few studies have considered the training level of a quantized LLM when evaluating their proposed low-bit quantization approaches.
    \item We derive scaling laws to model QiD with respect to the number of training tokens, model size and bit width. Using these scaling laws, we propose to use QiD as a signal to measure whether an LLM is fully trained and estimate the number of training tokens required for LLMs of different sizes to reach a fully trained state. Moreover, we use the scaling law to predict the performance of low-bit quantization for different-sized LLMs trained with 100 trillion tokens. Our projection indicates potential challenges for the future application of low-bit quantization.
    \item We release all the 1500+ quantized checkpoints used in this work to facilitate future research on this problem.
\end{itemize}

\section{Preliminary: Scaling Laws for Large Language Models}\label{sec:bg}

Scaling laws for large language models \citep{kaplan2020scaling,hoffmann2022training} are crucial for understanding how these models' performance improves with increased scale, including the number of parameters and training tokens:

\paragraph{Number of Parameters} LLMs' performance typically follows a power-law improvement as the number of parameters increases, allowing larger models to better fit and generalize on the same dataset:
\begin{equation}
    L(N) = \frac{a}{N^\alpha} + \epsilon
\end{equation}
where $L(N)$ is the loss function\footnote{We mainly discuss cross entropy loss for language modeling in this paper.} dependent on $N$ (the number of non-embedding parameters), $a$ is a constant (i.e., coefficient), $\alpha$ is the scaling exponent, and $\epsilon$ represents the error term. This relationship indicates larger models are generally more capable of capturing the complexities of language, leading to better generalization and lower loss.

\paragraph{Training Tokens} More training tokens also boost performance in a power-law fashion, enabling models to capture language complexities more effectively:
\begin{equation}
    L(D) = \frac{b}{D^\beta} + \epsilon
\end{equation}
where $D$ denotes the number of training tokens, $b$ is a constant (i.e., coefficient) and $\beta$ is the scaling exponent for training tokens. More training tokens enhance an LLM's ability to learn and generalize, allowing it to achieve better language modeling performance with lower loss.

When scaling both the number of parameters \( N \) and the amount of training data \( D \) simultaneously, the scaling law can be expressed as a function that accounts for the combined effects of both:
\begin{equation}\label{eq:oai_law}
    L(N, D) = [(\frac{N_c}{N})^{\frac{\alpha_N}{\alpha_D}} + \frac{D_c}{D}]^{\alpha_D}
\end{equation}
This scaling law allows us to estimate the performance of language models at unprecedented scales of model size and training data effectively before conducting actual training runs.

\section{Scaling Laws for Low-bit Quantization}\label{sec:scalinglaw}

In this section, we propose scaling laws for low-bit quantization. Unlike the scaling laws discussed in Section \ref{sec:bg}, the focus here is on understanding how quantization-induced degradation (QiD) changes when low-bit quantization is applied to LLMs of varying training scales. Formally, QiD is defined as follows:
\begin{equation}\label{eq:qid}
    \Delta_qLoss = Loss_{q} - Loss_{\textrm{16-bit}}
\end{equation}
where $Loss_q$ is the cross-entropy loss of a quantized LLM, and $Loss_{\textrm{16-bit}}$ is the cross-entropy loss of its pre-quantized counterpart with fp16 or bf16 weights. \(\Delta_qLoss\) represents QiD, which is the difference in loss before and after applying low-bit quantization.

Inspired by conventional scaling laws for language modeling, we investigate the impact of model size and the number of training tokens on QiD. Additionally, we consider bit width (i.e., the precision of quantized weight values).

\subsection{Experimental Setting}\label{subsec:setting}

We select open-sourced LLMs from the Pythia suite~\citep{biderman2023pythia} for our experiments. Pythia not only includes LLMs of various sizes, but also provides access to all checkpoints throughout its training process (from scratch to 300 billion tokens), allowing us to conduct experiments in a controlled setting to derive scaling laws for low-bit quantization.

We choose 6 different sizes of Pythia LLMs: 160M, 410M, 1B, 2.8B, 6.9B, and 12B. For each size, we sample 20 checkpoints (see Appendix \ref{sec:app_detail}) up to 98k steps\footnote{98k steps correspond to approximately 206 billion tokens, which is equivalent to one epoch of Pythia's training data. Although Pythia was trained for 143k steps, we skipped checkpoints beyond 98k steps to avoid the influence of duplicated data, as the data beyond 98k steps probably represents the second epoch with data that has already been trained with.}.

For quantization, we employ one of the most popular LLM quantization techniques -- GPTQ~\citep{frantar2022gptq} -- to quantize the Pythia checkpoints to 2-bit, 3-bit and 4-bit levels. 

We evaluate QiD on 1,000 randomly sampled texts from RefinedWeb dataset~\citep{penedo2023refinedweb}. 

\subsection{Training Tokens}\label{subsec:tokens}

In contrast to traditional language modeling scaling laws where the number of training tokens $D$ appears in the denominator, we propose the relationship between training tokens and QiD as follows:
\begin{equation}\label{eq:data}
    \Delta_qLoss(D) \approx b \cdot D^{\beta}
\end{equation}
because the more training tokens, the more significant the QiD becomes, according to our observations in Figure \ref{fig:intro_quant}.

\begin{figure}[ht]
\centering
    \includegraphics[width=1\textwidth{}]{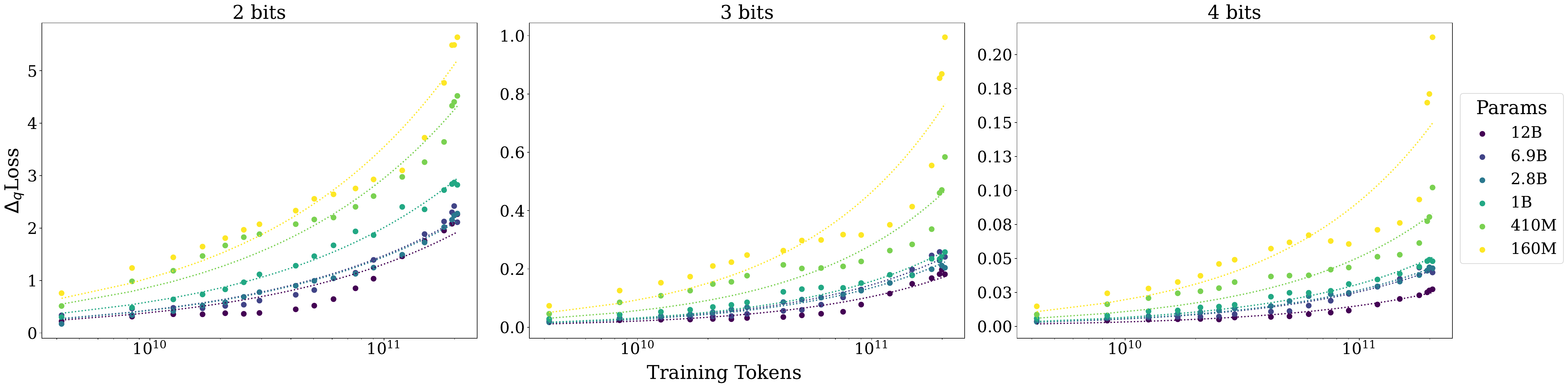}\vspace{-0.1cm}
    \caption{The fitted scaling law of QiD with respect to the number of training tokens in the form of Eq (\ref{eq:data}), where $\beta$ is fitted to be 0.5316.}\vspace{-0.2cm}
    \label{fig:scaling_tokens}
\end{figure}

We use the above functional form to fit the QiD observed in the quantized Pythia checkpoints in \ref{fig:scaling_tokens}, obtaining \(\beta=0.5316\), which fits the trend of QiD with respect to the change in training tokens quite well.

\subsection{Model Size}\label{subsec:param}

\begin{figure}[ht]
\centering
    \includegraphics[width=1\textwidth{}]{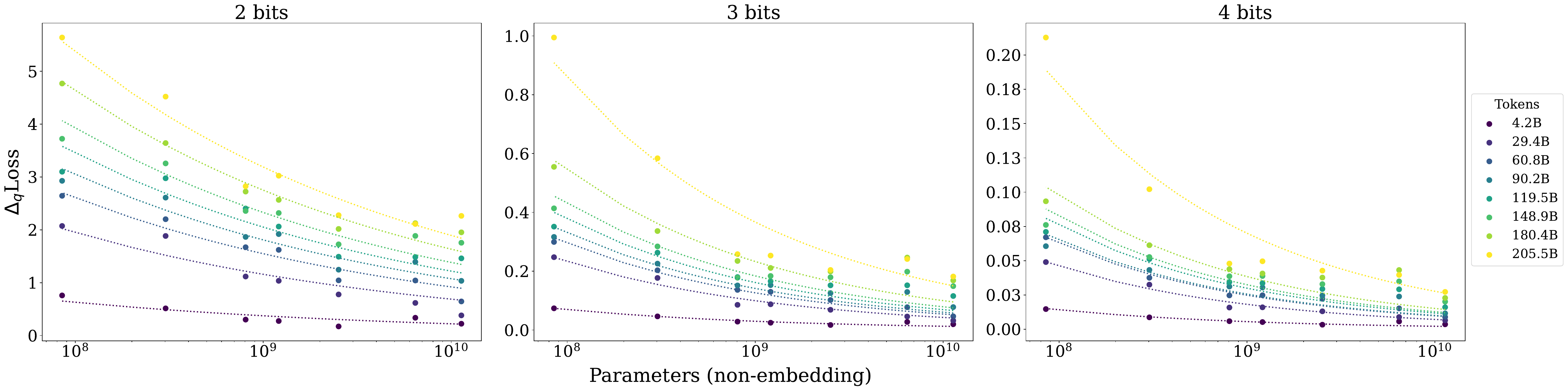}
    \caption{The fitted scaling law of QiD with respect to the model size (i.e., the number of non-embedding parameters) in the form of Eq (\ref{eq:params}), where $\alpha$ is fitted to be 0.2276.}\vspace{-0.2cm}
    \label{fig:scaling_params}
\end{figure}

As mentioned in Figure \ref{fig:intro_quant}, the larger the size of the model, the smaller the QiD tends to be. Therefore, we propose the relationship between model size (i.e., the number of non-embedding parameters) and QiD as follows:
\begin{equation}\label{eq:params}
    \Delta_qLoss(N) \approx \frac{a}{N^{\alpha}}
\end{equation}
We use the above functional form to fit the QiD of quantized Pythia checkpoints in Figure \ref{fig:scaling_params}, obtaining \(\alpha=0.2276\).

\begin{figure}[ht]
\centering
    \includegraphics[width=1\textwidth{}]{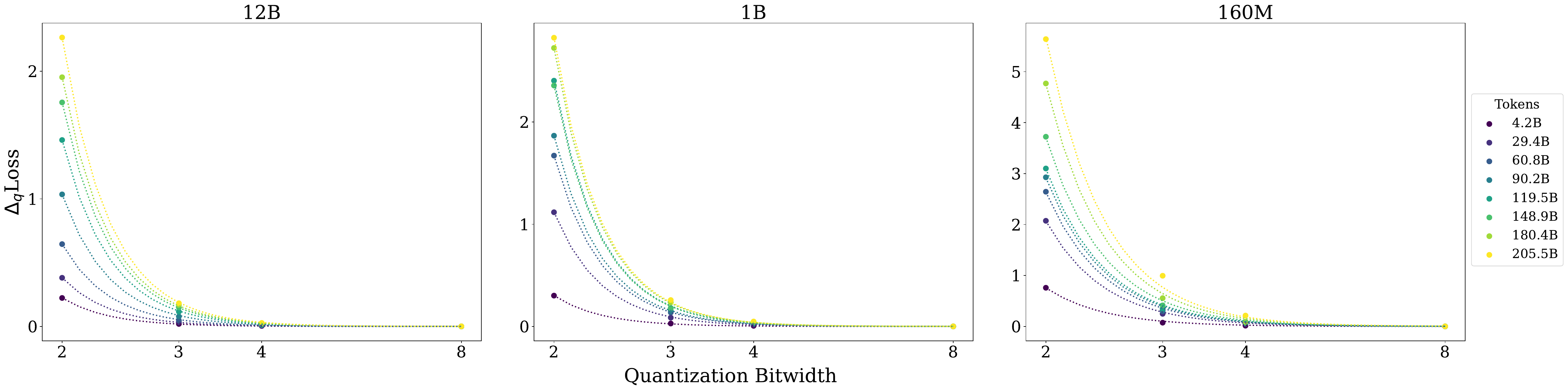}\vspace{-0.2cm}
    \caption{The fitted scaling law of QiD with respect to the bit width in the form of Eq (\ref{eq:bit}), where $\gamma$ is fitted to be 5.4812.}
    \label{fig:scaling_bit}
\end{figure}

\subsection{Bit Width}\label{subsec:bit}

Bit width is a factor not present in conventional scaling laws. Considering that the role of bit width is similar to that of the number of parameters (both aim to increase the model's expressiveness), we propose a similar functional form as in Section \ref{subsec:param} to model bit width in Eq (\ref{eq:bit}), and fit the data points of Pythia in Figure \ref{fig:scaling_bit}:
\begin{equation}\label{eq:bit}
    \Delta_qLoss(P) \approx \frac{c}{P^{\gamma}}
\end{equation}

\subsection{Unified Scaling Law}\label{subsec:unified_law}

With the basic scaling laws derived in Sections \ref{subsec:tokens} (the number of training tokens), \ref{subsec:param} (model size), and \ref{subsec:bit} (bit width), we study how to model QiD with all three factors together. Inspired by \citet{kaplan2020scaling}, we consider the following four principles for unifying the factors:

\begin{itemize}
\item Fixing D and P, sending \(N \to \infty\), we expect \(\Delta_q Loss \to 0\).
\item Fixing N and P, sending \(D \to 0\), we expect \(\Delta_q Loss \to 0\).
\item Fixing N and D, sending \(P \geq 16\), we expect \(\Delta_q Loss \to 0\).
\item Fixing N and D, sending \(P \to 0\), \(\Delta_q Loss\) should be very large.
\end{itemize}

\begin{figure}[ht]
\centering
    \includegraphics[width=1\textwidth{}]{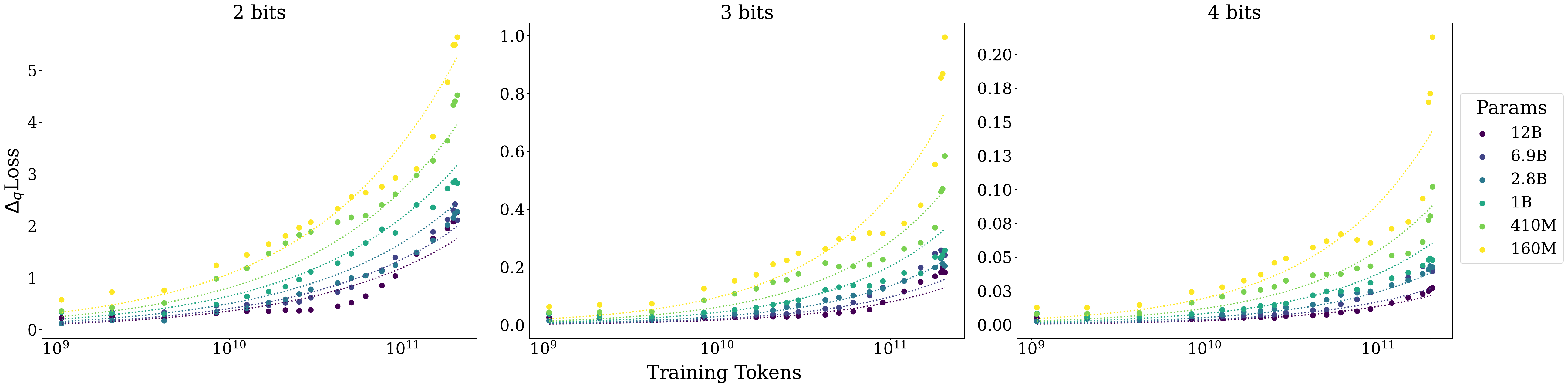}
    \caption{The unified scaling law we fit based on Eq (\ref{eq:joint}) with the GPTQ-quantized LLMs from the Pythia suite: $\Delta_qLoss(N,D,P)=0.017D^{0.5251} / (N^{0.2261} \cdot P^{5.4967})$}
    \label{fig:scaling_all_delta}
\end{figure}

We propose the unified scaling law for low-bit quantization as follows:
\begin{equation}\label{eq:joint}
    \Delta_qLoss(N, D, P) = k \cdot \frac{D^\beta}{N^\alpha P^\gamma}
\end{equation}
where \( k \) is the joint coefficient, and both the coefficient and exponents (\(\alpha\), \(\beta\), \(\gamma\)) are positive. Figure \ref{fig:scaling_all_delta} displays the fitted curves using this functional form. The jointly fitted exponents \(\alpha\), \(\beta\), and \(\gamma\) closely match those obtained by fitting these variables independently, further validating the effectiveness of the joint function form \( \Delta_qLoss(N, D, P) \).

\begin{figure}[ht]
\centering
    \includegraphics[width=1\textwidth{}]{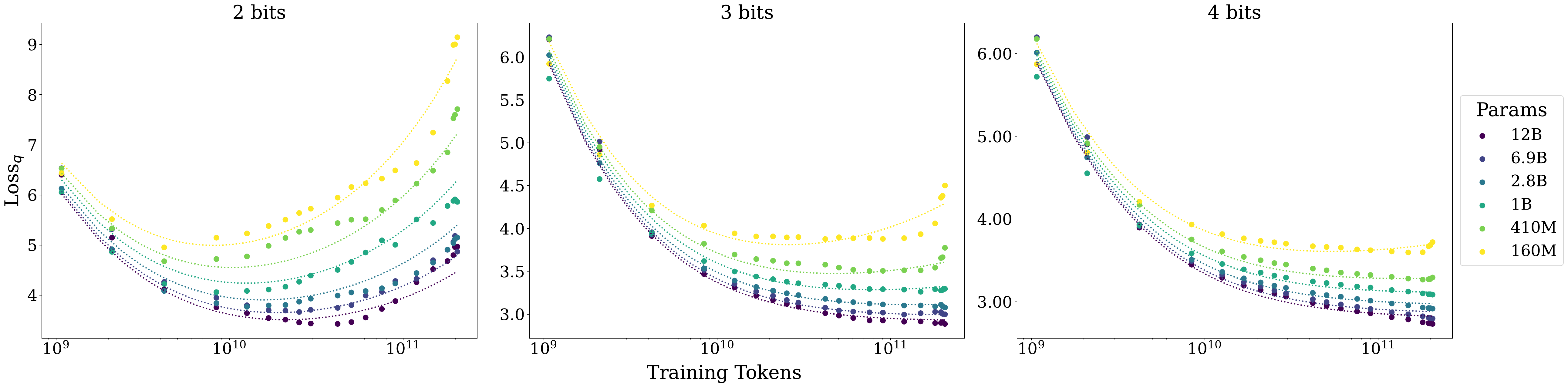}
    \caption{We can predict the performance of a quantized LLM as \(Loss_q = Loss_{\textrm{16-bit}} + \Delta_qLoss\), where $Loss_{\textrm{16-bit}}$ can be predicted by the conventional LLM's scaling law which is fitted based on the function form of Eq (\ref{eq:oai_law}) with the LLMs in the Pythia suite as $Loss_{\textrm{16-bit}}=[(4.74e^{19}/N)^{(0.045/0.399)}+7.63e^{10}/D]^{0.399}$.}
    \label{fig:scaling_all_loss}
\end{figure}

Given the unified scaling law for \(\Delta_qLoss\) and the definition of \(\Delta_qLoss\) in Eq (\ref{eq:qid}), we can easily predict a quantized LLM's performance as \(Loss_q = Loss_{\textrm{16-bit}} + \Delta_qLoss\), as illustrated in Figure \ref{fig:scaling_all_loss}, which fits well with the observed data points.

\subsection{Validation with Ablation Studies}

We validate the scaling law derived in Section \ref{subsec:unified_law} with different test data, quantization methods and foundation models.

\subsubsection{Test Data}

\begin{figure}[ht]
\centering
    \includegraphics[width=1\textwidth{}]{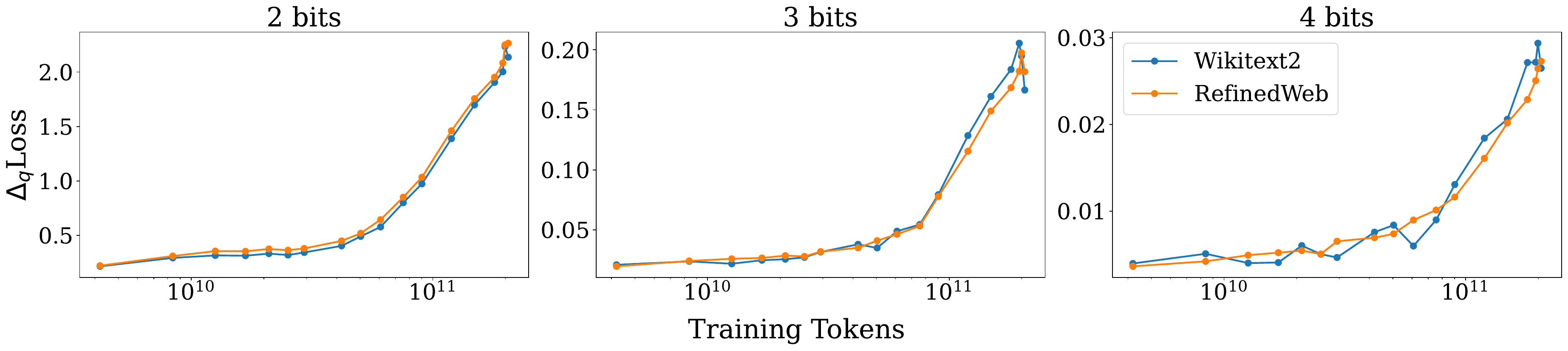}
    \caption{QiD results evaluated on RefinedWeb and Wikitext-2 with the 12B Pythia model.}
    \label{fig:ablation_data}
\end{figure}

We compare the results obtained using RefinedWeb and Wikitext-2~\citep{merity2016pointer} as test data in Figure \ref{fig:ablation_data}, demonstrating that the QiD results on these two test datasets are almost identical. This suggests that the trends of QiD are largely independent of the test data.

\subsubsection{Quantization Methods}

\begin{figure}[ht]
\centering
    \includegraphics[width=1\textwidth{}]{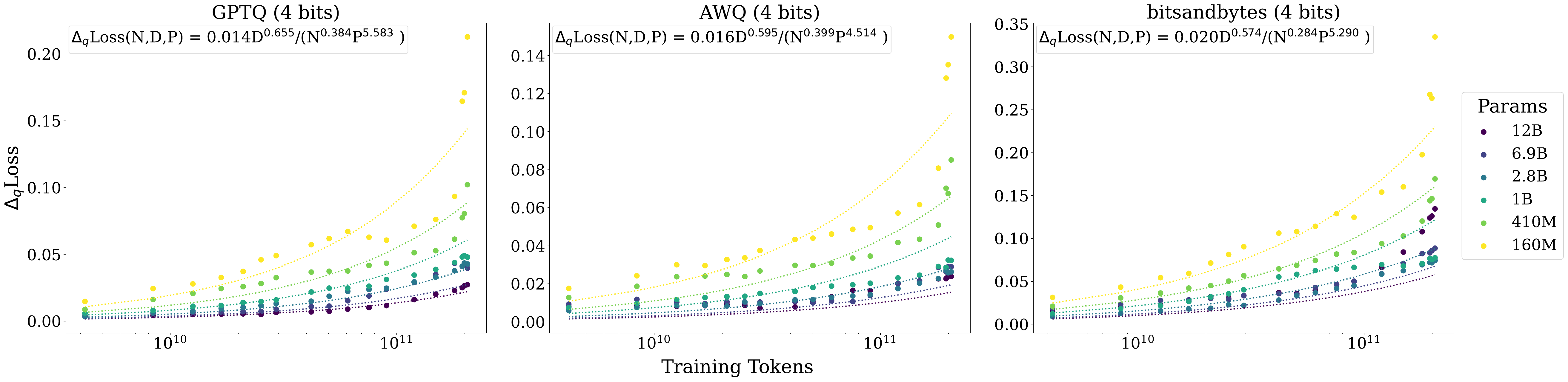}
    \caption{QiD results and fitted scaling laws for different quantization methods. Note that the GPTQ function here differs slightly from that in Figure \ref{fig:scaling_all_delta}, as it is fitted exclusively with 4-bit quantized Pythia checkpoints, whereas the function in Figure \ref{fig:scaling_all_delta} is fitted using all quantized Pythia checkpoints.}
    \label{fig:ablation_quant_methods}
\end{figure}

We quantize the Pythia checkpoints using another two popular quantization methods -- AWQ~\citep{lin2024awq} and bitandbytes\footnote{\url{https://github.com/bitsandbytes-foundation/bitsandbytes}} in addition to GPTQ. We show the QiD results and fitted scaling laws in Figure \ref{fig:ablation_quant_methods}, and we observe that the QiD trends for different quantization methods are almost identical, although the fitted scaling laws show slight differences.

\subsubsection{Foundation Models}

\begin{figure}[ht]
\centering
    \includegraphics[width=1\textwidth{}]{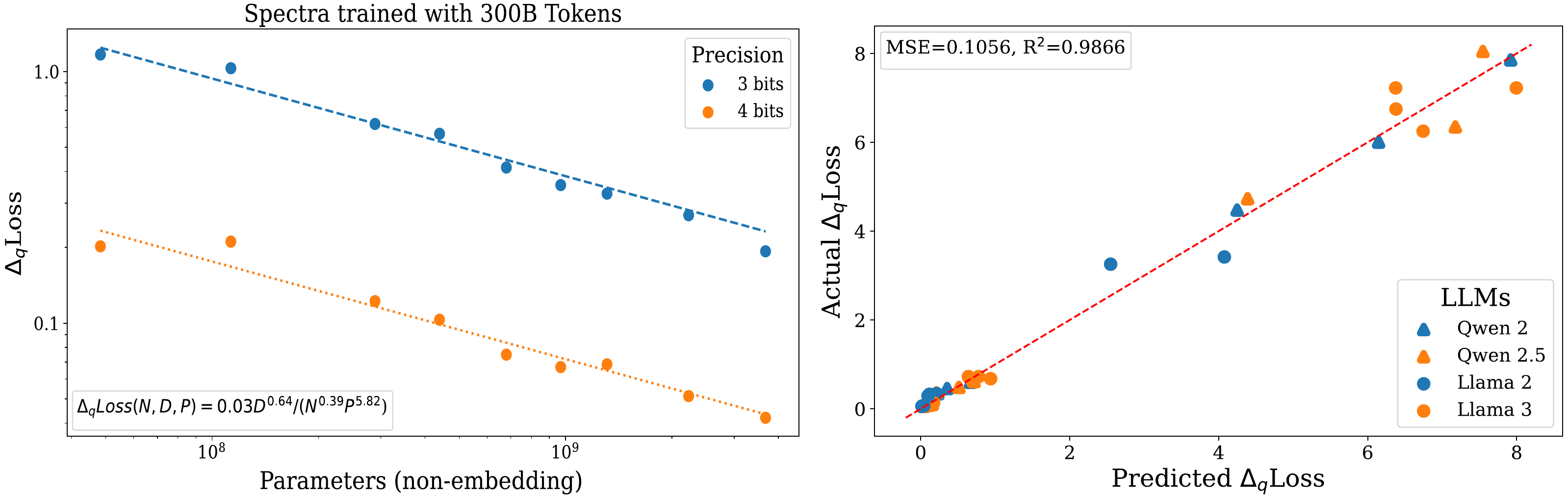}
    \caption{\textbf{Left: }Scaling laws for low-bit quantization, fitted on the LLM checkpoints of the Spectra suite, which are all trained with 300B tokens; \textbf{Right: }Actual $\Delta_qLoss$ \textbf{VS} Predicted $\Delta_qLoss$ that is computed based on the scaling laws fitted on Llama and Qwen.}
    \label{fig:ablation_models}
\end{figure}

Figure \ref{fig:ablation_models} shows the fitting results of our scaling laws function form, Eq (\ref{eq:joint}), on the Spectra suite~\citep{kaushal2024spectra} as well as the popular open-sourced Llama~\citep{touvron2023llama,dubey2024llama} and Qwen~\citep{yang2024qwen2} models, which confirms that the scaling laws are not only valid for Pythia but are likely to be broadly applicable.

\section{Discussion: Low-bit Quantization Favors Undertrained LLMs}
\subsection{Intuition}\label{subsec:intuition}

Based on the scaling laws we derived in Section \ref{sec:scalinglaw}, we  confirm low-bit quantization tends to favor models with fewer training tokens or larger model sizes, which are essentially  undertrained LLMs.

\begin{figure}[ht]
\centering
    \includegraphics[width=1\textwidth{}]{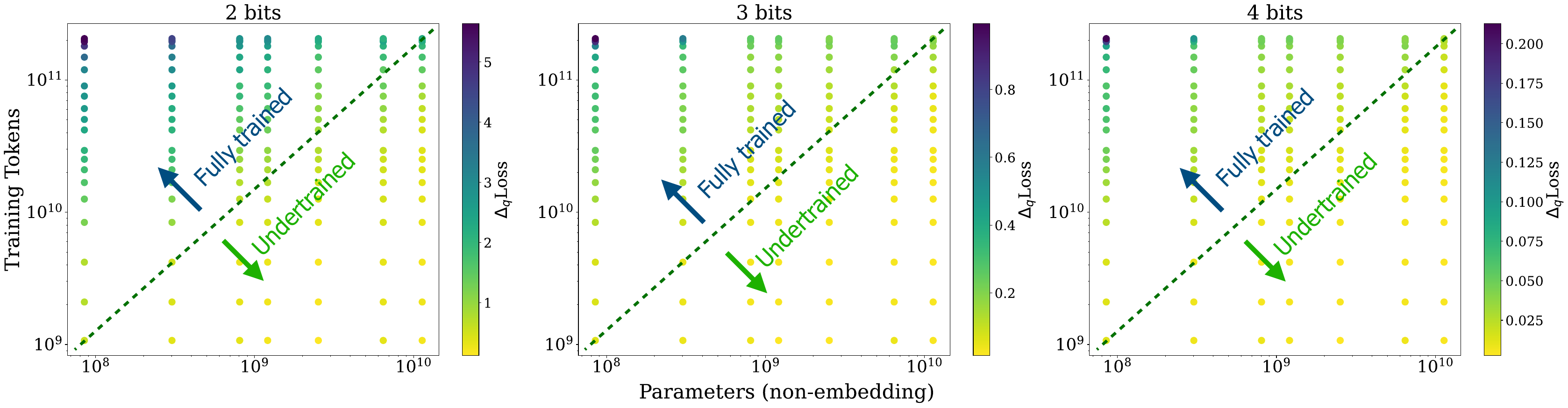}
    \caption{Fully trained LLMs suffer from much greater QiD (i.e., $\Delta_qLoss$) than undertrained LLMs.}
    \label{fig:scatter}
\end{figure}

Figure \ref{fig:scatter} illustrates the relationship between QiD, model size, and training tokens. Points located in the upper-left corner are more fully trained and have a much higher QiD, while points in the bottom-right corner are more undertrained and have a lower QiD.

\begin{figure}[ht]
\centering
    \includegraphics[width=0.95\textwidth{}]{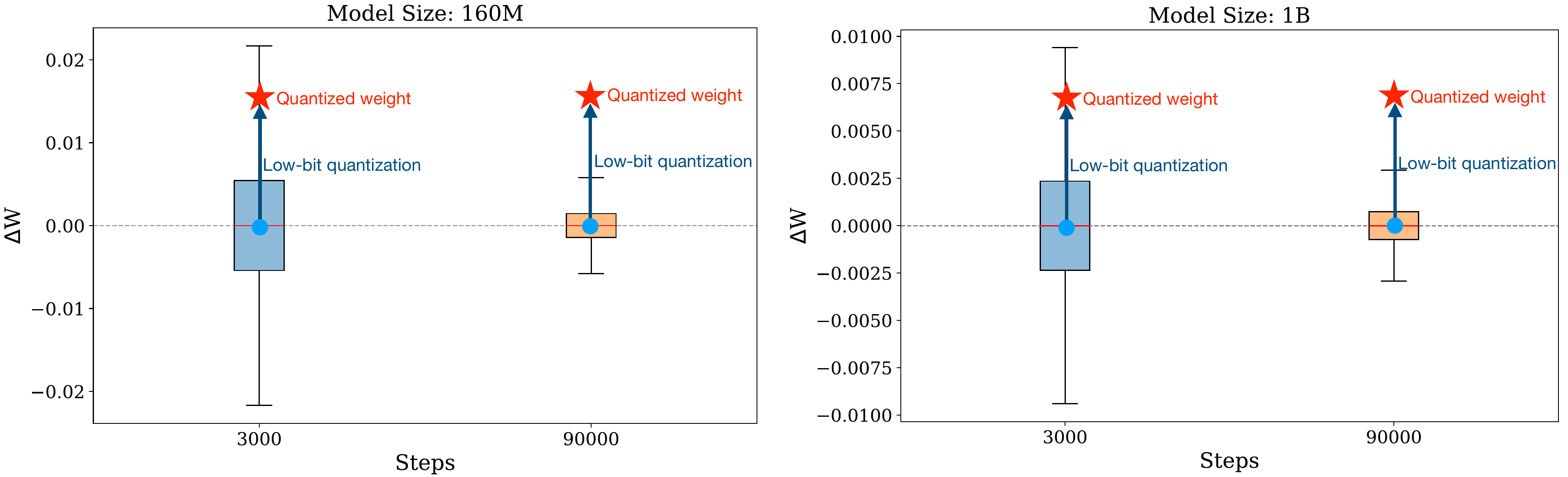}
    \caption{Changes in model weights between adjacent checkpoints. Early (undertrained) checkpoints exhibit significant weight fluctuations during training, making the model relatively robust to weight variations. Therefore, small changes introduced by quantization have a limited impact on the model's performance. In contrast, fully trained checkpoints demonstrate very little weight fluctuations during training. As a result, low-bit quantization is likely to push weights \textbf{beyond the narrow range of recent variations}, leading to performance degradation or even model collapse.}
    \label{fig:quant_bound}
\end{figure}

To understand this observation intuitively, we illustrate changes in sampled model weights between adjacent checkpoints in Figure \ref{fig:quant_bound}. It can be observed that the early checkpoints exhibit significant changes in weights. Due to the significant fluctuations in weights during training, the model becomes inherently robust to weight variations, meaning that even if low-bit quantization introduces some precision loss, the overall impact on the model remains limited. On the other hand, checkpoints from the later stages of training, which are more fully trained, show very small changes in weights (often at a very small scale, even beyond the 3rd-4th decimal place). In such cases, low-bit quantization is very likely to shift weights outside the small range of recent variations, potentially causing the model to degrade or even collapse.

From another perspective, during the undertrained stage, the model's weights undergo significant changes and have not yet fully exploited the precision dimension. In the later, more fully trained stage, as weight adjustments stabilize, the model increasingly relies on precision to continue optimizing the training objective and improving language modeling performance. This aligns with the two phrases of representation learning in the information bottleneck theory~\citep{shwartz2017opening}: during the early training phase, gradients have a large mean and small variance, making high precision unnecessary. However, in the later training phase, gradients have a small mean and large variance, requiring higher precision for the model to converge effectively.

\subsection{QiD: A Signal that Measures an LLM's Training Level}

Unlike previous work that often uses the inability of the loss to decrease further as a signal to determine whether an LLM is fully trained (i.e., saturated), we introduce a novel perspective that we can use QiD to determine whether an LLM is fully trained. If an LLM exhibits QiD $\approx$ 0 after low-bit quantization, it suggests that the LLM is likely undertrained, as it has not yet exploited higher precision, as discussed in Section \ref{subsec:intuition}.

\begin{table}[t]
\centering
\caption{Prediction of the number of training tokens (in trillion) needed to achieve a given training level measured by \(\Delta_qLoss\) for different model sizes and bit widths. Note that $\Delta_qLoss=0.2$ means the likelihood is reduced to 80\% of its original value ($e^{-0.2} \approx 0.8$), while $\Delta_qLoss = 0.5$ means the likelihood is reduced to 60\% ($e^{-0.5} \approx 0.6$).}\label{tab:deltapredict}
\vspace{0.2cm}
\resizebox{\textwidth}{!}{%
\begin{tabular}{c|ccc|ccc|ccc|ccc@{}}
\toprule
\multirow{2}{*}{Model Size} & \multicolumn{3}{c|}{$\Delta_q$Loss = 0.2} & \multicolumn{3}{c|}{$\Delta_q$Loss = 0.3} & \multicolumn{3}{c|}{$\Delta_q$Loss = 0.4} & \multicolumn{3}{c}{$\Delta_q$Loss = 0.5} \\ 
            & 2 bits         & 3 bits        & 4 bits         & 2 bits         & 3 bits        & 4 bits         & 2 bits        & 3 bits        & 4 bits          & 2 bits        & 3 bits        & 4 bits         \\ \midrule
1B                      & 0.0011       & 0.1089      & 1.4424       & 0.0025       & 0.1990      & 2.6786       & 0.0043      & 0.3051      & 4.1556        & 0.0066      & 0.4251      & 5.8422       \\ \midrule
7B                  & 0.0026       & 0.3038      & 4.5066       & 0.0057       & 0.5550      & 8.3689       & 0.0099      & 0.8512      & 12.9836       & 0.0152      & 1.1860      & 18.2531      \\ \midrule
70B                    & 0.0071       & 1.0228      & \bf 17.3499      & 0.0154       & 1.8687      & 32.2192      & 0.0267      & 2.8659      & 49.9854       & 0.0409      & 3.9932      & 70.2723      \\ \midrule
405B                   & 0.0151       & 2.5807      & \bf 48.4861      & 0.0328       & 4.7151      & 90.0398      & 0.0567      & 7.2311      & 139.6892      & 0.0868      & 10.0754     & 196.3829     \\ \bottomrule
\end{tabular}%
}
\end{table}

With the scaling law in Eq (\ref{eq:joint}) derived in Section \ref{subsec:unified_law}, we can estimate how many training tokens are needed for a given LLM size to be considered fully trained based on QiD predictions. Table \ref{tab:deltapredict} shows the number of training tokens required for different model sizes to achieve $\Delta_qLoss$ = \{0.2, 0.3, 0.4, 0.5\} when applying low-bit quantization. For a 70B scale model, achieving a QiD greater than 0.2 (corresponding to likelihood decrease by 20\%) under 4-bit quantization requires over 17 trillion training tokens. In contrast, for a 405B scale LLM, achieving a QiD above 0.2 under 4-bit quantization requires nearly 50 trillion training tokens -- a scale far beyond what has been achieved by now, indicating that current training efforts for extremely large LLMs may be still far from sufficient.

\begin{figure}[ht]
\centering
    \includegraphics[width=0.94\textwidth{}]{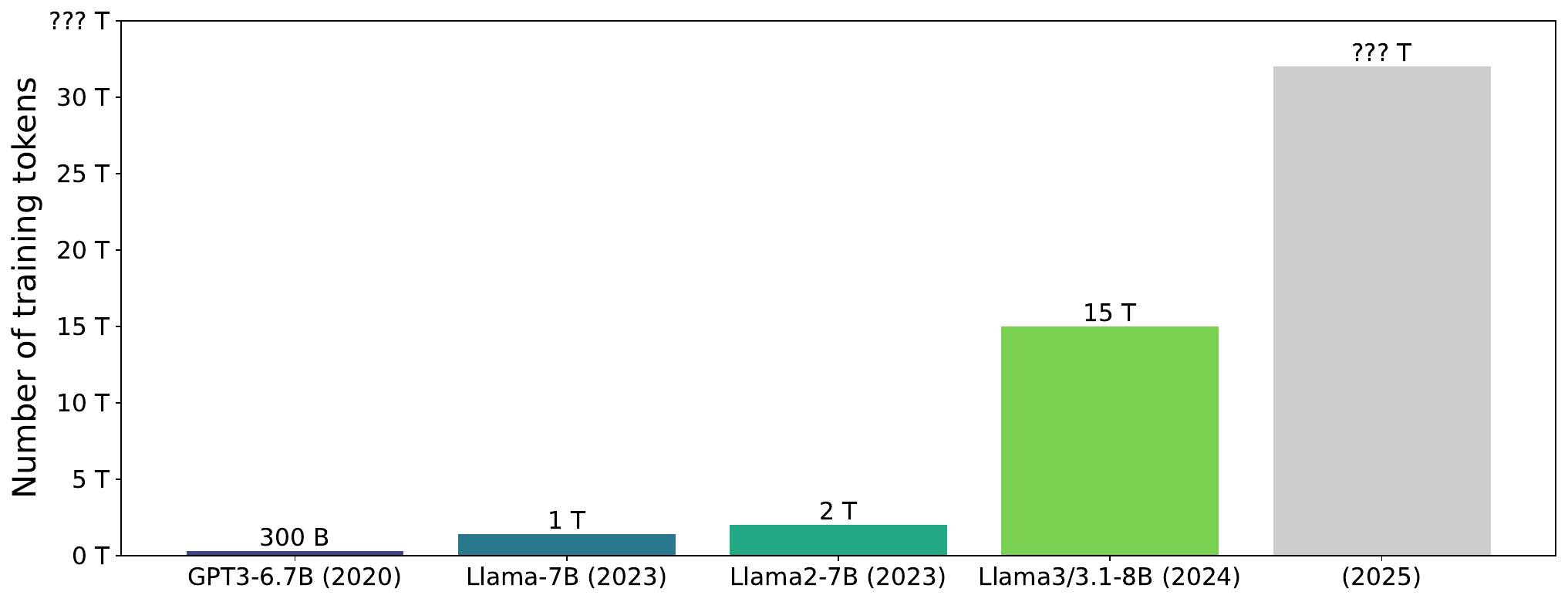}
    \caption{The number of training tokens for the state-of-the-art 7B-scale LLMs increase by nearly $50\times$ over the past 4 years. According to this trend, it is expected that the future models will have much more training tokens.}
    \label{fig:7bllms}
\end{figure}

\subsection{QiD Prediction When Scaling to 100 Trillion Training Tokens}

Figure \ref{fig:7bllms} shows the trend in the number of training tokens for state-of-the-art 7B-scale LLMs from 2020 to the present, showing that the number of training tokens has increased nearly $50\times$ over the past 4 years. Based on this trend, it is very likely that LLMs in 2025-2026 will be trained with up to 100 trillion ($10^{14}$) tokens\footnote{Although there have been claims that internet data is nearing exhaustion, recent continuous innovations in synthetic data creation~\citep{ge2024scaling} lead us to believe that the milestone of 100 trillion training tokens is achievable in the next few years.}.

Using the scaling laws derived, we predict the performance of quantized LLMs trained on 100 trillion tokens, as illustrated in Figure \ref{fig:100T} at the beginning of this paper. In particular, performance degradation with 2-bit and 3-bit quantization at the unprecedented training scale of 100 trillion tokens is predicted to be severe, which is in stark contrast to the acceptable performance at the current training scale of \(10^{13}\) tokens. This indicates a challenge for the practical application of low-bit quantization to future LLMs.

\subsection{From Low-bit Quantization to Low-bit LLMs}

\begin{figure}[h!]
    \centering
    \captionsetup{skip=-5pt}
    \begin{subfigure}[b]{0.49\textwidth}
        \centering
        \includegraphics[width=\textwidth{}]{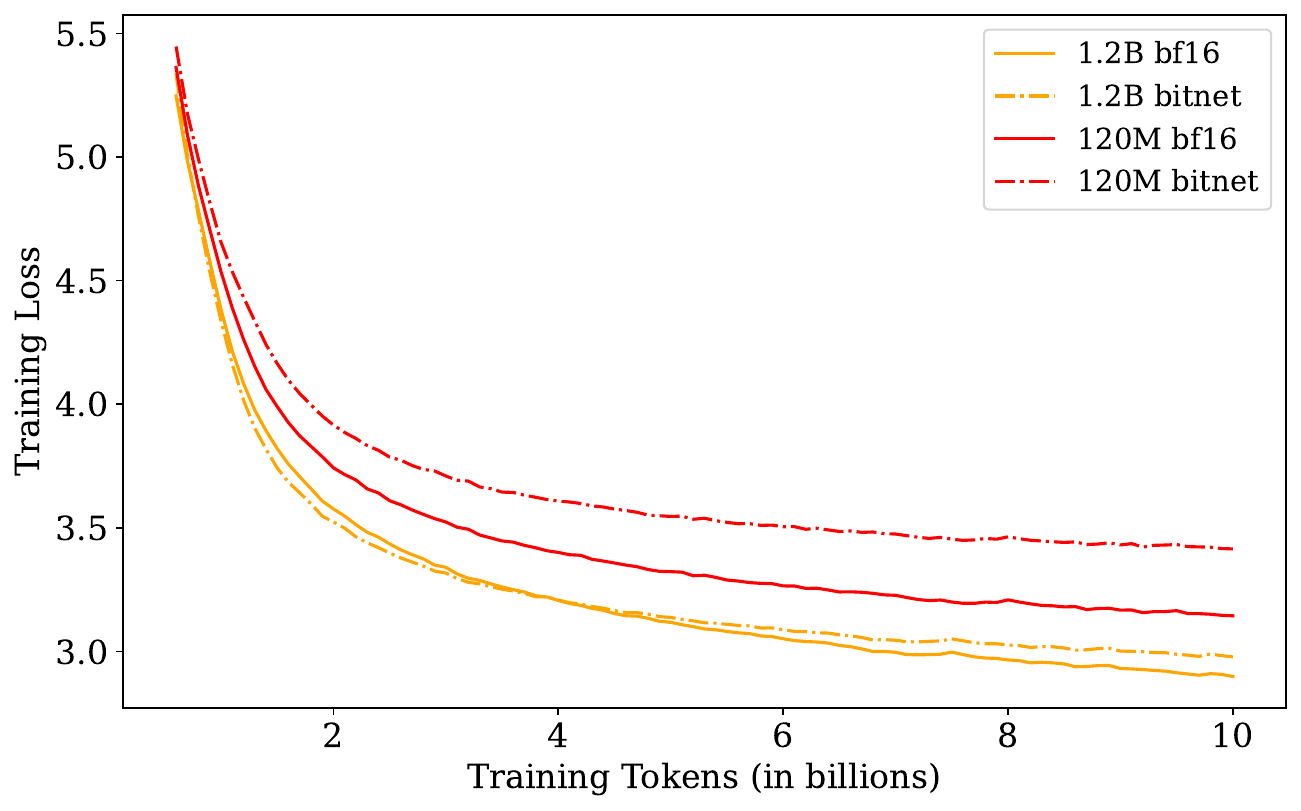}
        \label{fig:early}
    \end{subfigure}
    \begin{subfigure}[b]{0.49\textwidth}
        \centering
        \includegraphics[width=\textwidth{}]{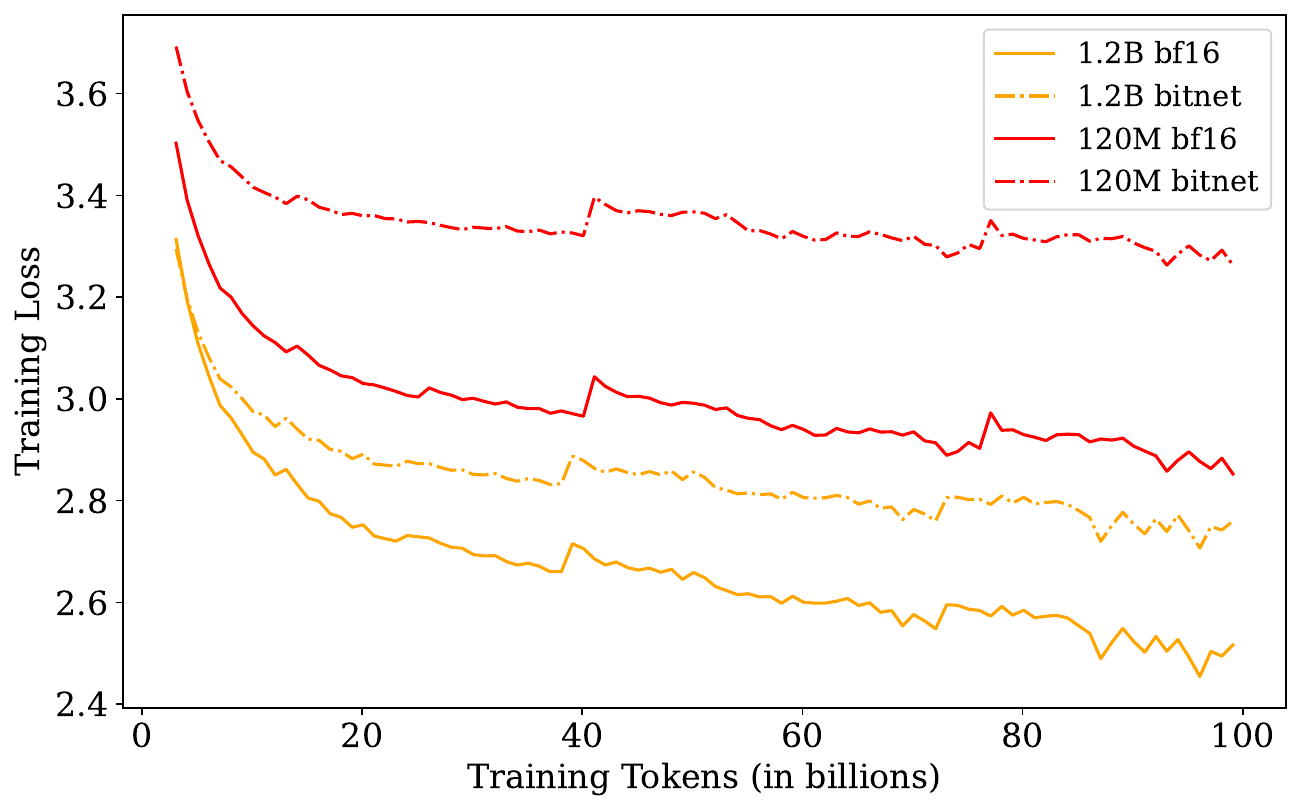}
        \label{fig:mid}
    \end{subfigure}
    \caption{Training losses of BitNet and its 16-bit counterparts show a trend similar to that of low-bit quantization -- they tend to perform well when undertrained but struggle to match the performance of fully trained LLMs.}
    \label{fig:bitnet}
\end{figure}

Although this work mainly focuses on low-bit (post-)quantization, we suspect that native low-bit LLMs are also likely to favor undertrained LLMs. We replicated the popular 1-bit LLM -- \textit{BitNet b1.58}~\citep{ma2024era} -- to compare it with its bf16 counterpart throughout training. Specifically, we trained 120M and 1.2B decoder-only models with both bf16 and BitNet. Figure \ref{fig:bitnet} shows the comparison of training losses between BitNet and its 16-bit counterparts in the early- and mid-training steps. It can be observed that, in the early stages of training, the training loss curves of BitNet closely match (and even outperform) those of bf16, as BitNet tends to use a higher learning rate than bf16 training according to its training recipe. As training continues, the 120M BitNet gradually begins to lag behind its bf16 counterpart, and after further training steps, a noticeable gap starts to appear in the 1.2B models, which is consistent with our observations for low-bit quantization. This indicates that native low-bit LLMs such as BitNet\footnote{We reviewed the original BitNet paper and some open-sourced reimplementations, and found that their numbers of training tokens were up to 100 billion. Considering their model sizes and the fact that the performance gap of native low-bit LLMs tends to emerge later compared to post-quantization, we express concerns about their performance at larger training scales (i.e., with more training tokens). We call for results of native low-bit LLMs at larger training scales to better justify their practical value.} may also favor undertrained LLMs, though the gap manifests later compared to post-quantization, as native low-bit training keeps the model capable of operating under low precision throughout the training process.

\section{Conclusion}

We derive scaling laws for low-bit quantization from over 1500 quantized LLM checkpoints, and reveal that low-bit quantization favors undertrained LLMs. We provide an intuitive interpretation for this phenomenon and introduce a novel perspective of using QiD as a signal to determine a model's training level. Moreover, we use the derived scaling laws to predict the effect of low-bit quantization on LLMs trained with 100 trillion tokens. This, on one hand, challenges the future practical value of low-bit quantization, and on the other hand, suggests that future research on low-bit quantization should consider the model's training level during evaluation. Alongside concurrent research~\citep{kumar2024scaling,feng2024numerical} that takes a serious look at the limits of low-bit LLMs, we hope this work can help the community cool down from the surrounding hype, and foster deeper reflection and critical examination in this field.

\section*{Limitations}

This work includes the following limitations:

\begin{itemize}[left=1pt]
\item Although we have done our best to conduct extensive experiments and derive the scaling laws from over 1500 quantized checkpoints, it is still not extensive enough. For example, the training tokens used in our experiments with Pythia only amount to 300 billion. We expect more observations from a greater number of quantized checkpoints in the future to refine the scaling laws we have derived.
\item The scaling laws derived in this work are primarily focused on single-stage pre-trained language models. However, advanced LLMs today often employ multi-stage training strategies including supervised fine-tuning and preference optimization, and even within pre-training, multiple stages are often involved (e.g., Llama-3.1 focuses more on high-quality text, math, reasoning, and code data during the final pre-training stages). Such multi-stage training strategies may cause the behavior of the model after quantization to be significantly different, which we plan to explore in future work.
\end{itemize}

\bibliography{references}
\bibliographystyle{references}


\appendix

\section{Appendix}

\subsection{Implementation Details}\label{sec:app_detail}
\paragraph{Checkpoints of the Pythia}We choose the following 20 checkpoints of the Pythia models at the following steps for fitting the scaling laws: \{512, 1k, 2k, 4k, 6k, 8k, 10k, 12k, 14k, 20k, 24k, 29k, 36k, 43k, 57k, 71k, 86k, 93k, 95k, 98k\}.

\paragraph{Tokenization consistency}To ensure consistency in token counts for computing cross entropy loss, which can vary with different tokenizers, we use the token counts generated by the Llama-3 8B~\citep{dubey2024llama} tokenizer for all QiD calculations in this work.

\end{document}